\documentclass[conference]{IEEEtran}
\usepackage{cite}
\usepackage{amsmath,amssymb,amsfonts}
\usepackage{algorithmic}
\usepackage{graphicx}
\usepackage{textcomp}
\usepackage{xcolor}
\usepackage{url}
\def\BibTeX{{\rm B\kern-.05em{\sc i\kern-.025em b}\kern-.08em
    T\kern-.1667em\lower.7ex\hbox{E}\kern-.125emX}}
\begin{document}

\title{Training Electric Vehicle Charging Controllers with Imitation Learning}

\author{\IEEEauthorblockN{Martin Pil\'at}
\IEEEauthorblockA{\textit{Charles University, Faculty of Mathematics and Physics} \\ Prague, Czech Republic \\
Martin.Pilat@mff.cuni.cz, ORCID: 0000-0003-1239-1566}
}

\maketitle

\begin{abstract}
The problem of coordinating the charging of electric vehicles gains more importance as the number of such vehicles grows. In this paper, we develop a method for the training of controllers for the coordination of EV charging. In contrast to most existing works on this topic, we require the controllers to preserve the privacy of the users, therefore we do not allow any communication from the controller to any third party.

In order to train the controllers, we use the idea of imitation learning -- we first find an optimum solution for a relaxed version of the problem using quadratic optimization and then train the controllers to imitate this solution. We also investigate the effects of regularization of the optimum solution on the performance of the controllers. The method is evaluated on realistic data and shows improved performance and training speed compared to similar controllers trained using evolutionary algorithms.
\end{abstract}
    
\begin{IEEEkeywords}
Imitation learning; apprenticeship learning; neural networks; electric vehicle charging
\end{IEEEkeywords}

\section{Introduction and Related Work}

The problem of the coordination of electric vehicle (EV) charging is gaining more attention in recent years as the number of EVs rises, leading to increased importance of such coordination. Without coordination the charging can cause large peaks in the electricity consumption as many people have similar daily schedules -- they return home from work at similar times in the afternoon and if everyone started charging their EV immediately after they get home, the additional load might cause problems in the grid.

The goal of the coordination is to find an algorithm that would lower such peaks and fill the nightly valleys in the electricity consumption. Such algorithms have been developed in recent years, however many of these require centralized planning~\cite{4839973}. Such algorithms can have difficulties to scale with the number of vehicles. These problems can be partially solved by using pre-computed schedules together with their real-time adaptation~\cite{8703849}, or by de-centralization of the control, in the sense that at least some of the computation is performed by the controllers themselves instead of centrally. The de-centralized approaches are often based on iterative protocols, where the grid provides some charging signal. Such a protocol was developed for example by Gan \emph{et al.}~\cite{gan2013optimal}, they even show that under some assumptions (like knowing all the charging requests beforehand) the protocol is optimal. For more examples of algorithms for the coordination of EV charging, we recommend the recent survey paper by Al-Ogaili \emph{et al.}~\cite{8825773}.

Most of the currently existing controllers require some communication with a third party in order to coordinate the charging. Some consumers may not want to share their charging requests with anyone, as the request indicates when the customers are at home and when they plan to leave (when the request ends). We considered this problem in our previous work~\cite{8489027}, where we developed de-centralized controllers that preserve this privacy. The controllers require no external communication or only a one-way one (from grid to controller). The charging requests are never shared with a third party. Customer privacy is also one of the open problems mentioned in the above survey paper. The main disadvantage of the controllers presented in our previous paper is that their training is quite time-consuming as it is done by an evolutionary algorithm that needs to simulate the grid in each fitness evaluation. 

In this paper, we approach the problem of the training from another angle -- by using the idea of imitation learning (also known as apprenticeship learning), where the goal is to learn behavior by observing an expert. Similar methods were used already almost 30 years ago to train autonomous navigation~\cite{pomerleau1991efficient} and are still commonly used and studied with applications in control tasks~\cite{NIPS2016_6391}. In this paper, the controllers are trained to imitate on optimal schedule found by solving a relaxed version of the problem -- in the relaxation we assume we know all the charging requests and all future consumptions. Solution of this problem is easy and fast to obtain. The controllers themselves are then trained by supervised learning with a training set created from the optimum solution. While the optimum solution is computed with the knowledge of the future, the controllers do not need any inputs that would not be available at the time they make the decision. 

The use of imitation learning for the control of EV charging is the main contribution of this paper compared to our previous work. Apart from that we also investigate, if the solutions could be regularized in such a way that they would be easier to train for the controllers, without reducing their performance. We show that the resulting controllers can be obtained faster than by using evolutionary algorithms and that they even provide better results. 

Compared to other literature on EV charging, the main advantage of the controllers presented here is the fact that they preserve the privacy of the users -- we are not aware of any other controllers that would have this feature.

\section{Charging Coordination by Imitation}

We assume each household has a controller (smart charger) and when the user gets home, they connect their EV to the charger and specify the time when the car needs to be fully charged. The controller reads the information about how much charge is needed by the EV and fulfills the charging request by specifying, in each time step, the charging speed to use in the following time step.

In this section, we describe the controllers in detail. We start with the description of the inputs available to the controllers, and how they are processed to create the features that are directly used to set the charging speed. These features are mostly the same as in our previous work~\cite{8489027}, but we include them here for completeness. Further details on the model and simulator can also be found in that paper.

Then, we describe the relaxed version of the problem that can be solved using quadratic optimization and show, how the result can be used to create a training set and train the controllers.

\subsection{Controller Inputs}

As, for privacy reasons, we only allow one way communication from the grid to the controller the available inputs are limited. We define two types (A and H) of controllers. Controllers of the H-type use no external communication and their only inputs are from the user (the charging request) and the current electricity consumption of their household (measured locally). The controllers of the A-type additionally receive the information about the total consumption of all households -- here we assume there is an electricity meter for the whole neighborhood that provides this information with enough granularity.

More specifically, the inputs for the H-type controllers are 
\begin{enumerate}
    \item current time and date,
    \item information about the current charging request (if any) -- the needed charge, maximum possible charging speed, and the time, when the request must be finished, and
    \item the current consumption of the household.
\end{enumerate}

The A-type controllers additionally receive the current consumption of the whole grid as one of the inputs.

The information about the charging request is available only in those time steps a charging request is active for the given controller (i.e. a car is available and is not fully charged). In the other time steps, only the consumption information and current date and time is available. The controllers need to provide an output only if a charging request is active.

The controller processes the information given in each time step to create features related to the recent consumption in the household/the whole grid. The features for household consumption and the grid consumption are computed in the same way. More specifically, we use the following features:
\begin{enumerate}
    \item the relative change of the consumption since the last step, last hour, and last three hours computed as $\Delta c_{t,p} = c_t/c_{t-p} - 1$, where $c_t$ is the consumption at time $t$ and $p$ is the duration of one step, 1 hour or 3 hours;
    \item the current consumption linearly scaled between 0 and 1, where 0 is the minimum and 1 is the maximum consumption over the last 24 hours;
    \item the current consumption divided by the average consumption in the last 24 hours.
\end{enumerate}

Altogether, we have five different features regarding the household consumption and its changes and, for the A-type controllers, additional five features regarding the whole grid.

Apart from these features, we also compute additional features whenever an active charging request is available, these are:
\begin{enumerate}
    \item the remaining percentage of the original needed charge, 
    \item the remaining percentage of the original time steps,
    \item the minimum required charging speed needed in the current time step to finish the charging in time,
    \item the charging speed that would finish the charging in time, if it was used for all remaining time steps,
    \item the length of the request in days,
    \item the current time and the ending time of the request encoded as a pair $(\sin(2\pi t/86400),\cos(2\pi t/86400))$ where $t$ is the time since the last midnight in seconds, and
    \item an indicator variable that is 0 on workdays and 1 on weekends.
\end{enumerate}

The charging speeds are normalized to the interval $[0,1]$ by division by the maximum charging speed (that is given by the EV). The request length is in days in order to normalize the values between 0 and 1 (most request are shorter than 24 hours), and the time encoding ensures continuous changes throughout the day. Overall, we have ten features related to the charging request. 

\subsection{The controller}

The controller can be generally described as a function $\mathcal{C}(f, \theta)$ with internal (trainable) parameters $\theta$ mapping the features $f$ described above to the interval $[0,1]$. The particular representation of the function can vary, however in this work we assume the function is represented by a neural network with weights $\theta$, specifically, we experiment with small networks with a single hidden layer with 5 or 50 neurons (cf. Section~\ref{sec:experiments}). The output of the function is interpreted as the percentage of the maximum charging speed that should be used in the next time step, i.e. the charging speed for the next time step is $\mathcal{C}(f, \theta) \cdot m$, where $m$ is the maximum charging speed.

Taking the outputs of the controller function directly might mean that the requests are not fully satisfied in time or that the battery would overcharge. Therefore, we augment the returned value by making sure it is always between the minimum required charging speed and the speed that would fully charge the battery in the current time step (if it is lower than the maximum). This simple transformation of the outputs ensures that all the requests are always satisfied, if it is possible to satisfy them.

The goal of the controller training is to set the parameters $\theta$ in such a way that the overall consumption of the grid (obtained by simulation) matches some desired function. In this paper, we try to minimize the standard deviation of the consumptions in time, which essentially means that we try to match a constant function (the sum of the baseline uncontrollable consumptions and the charging requests is constant, therefore minimizing the standard deviation is the same as minimizing the squares of the total consumptions). The problem of setting the parameters $\theta$ can be expressed as a reinforcement learning one, however, it is quite complex, as we try to learn cooperation of the controllers without any direct communication between them and without the knowledge of the future consumptions and requests. In previous work~\cite{8489027}, we used evolutionary algorithm and gradient descend in order to optimize the parameters $\theta$. Here, we propose to use imitation learning to set the parameters -- we find an optimum solution to a relaxed version of the problem and then use this solution to create a training set for the controllers. The controllers themselves can then be trained with any supervised learning technique.

\subsection{Generating the Training Set}
\label{sec:quad_opt} 

In order to create the training set, we first describe the problem of EV charging coordination as a quadratic optimization problem, with the goal to minimize the standard deviation of the electricity consumption in the grid and with constraints that ensure all the charging requests are met and the charging speed is always below the maximum allowed for a given request. The corresponding quadratic optimization problem is thus
\begin{equation*}
    \begin{aligned}
    & {\text{min}} 	     & & \mathrm{var_t}(B_t + \sum_{r \in \mathcal{R}\, s.t.\, t\in A(r)} r_t) \\
    & \text{subject to}  & & \sum_{t \in A(r)} r_t = R(r), \forall r \in \mathcal{R} \\
    & 					 & & 0 \leq r_t \leq M(r), \forall r \in \mathcal{R}, t \in A(r)\\
    \end{aligned}
\end{equation*}
where, $\mathcal{R}$ is the set of all charging requests, $A(r), R(r)$, and $M(r)$ are the times when the request $r$ is active, the total charge required by the request $r$ and the maximum charging speed of the request $r$ respectively. $B_t$ is the baseline uncontrollable consumption of the grid in time $t$ and $r_t$ is the charging speed for request $r$ at time $t$. 

In the formalization, we used the knowledge of all charging requests and electricity consumption in some time period -- we assume this is available for some time in the past to create the controllers. 

The optimization problem, although quite large for larger number of households with controllers and longer time periods can be easily solved using existing quadratic optimization solvers (in our experiments we solved the problem for 50 households simulated over the period of three months in a matter of minutes).

The solution to this problem gives us, for each charging request $r$ and each time step $t$ the charging speed $r_t$ that should have been used in order to minimize the variance of the electricity consumption of the grid. Based on these solutions, we can create a training set $T$, where we compute in each timestep and for each household the features described above and, if the household has an active request $r$ set the charging speed $r_t$ as a target for the controller $\mathcal{C}(f, \theta)$ (if there is no charging request active for the given household at time $t$, nothing is added to the training set). The training of the controller is then a simple regression problem that can be solved using existing methods. For example, if the controller is represented as a neural network, we can set its weights ($\theta$) using stochastic gradient descend.

\subsection{Stabilization of the Schedule}

A potential problem with the solution to the quadratic problem above may be that it is not unique, and specifically, if we have two charging requests active in two consecutive time steps, there are many different charging speeds each of them can use and still obtain the same state after the two time steps. For example, both can charge with half of the maximum speed in both time steps, or one of them can use the maximum speed in the first step and no charging in the second one, while the other does not charge in the first step and uses maximum speed in the second one. Both these solutions would have exactly the same objective and which one is found by the solver depends mostly on its implementation details. Moreover, we would expect that the former one would be much easier to imitate by the controller that the latter one as the outputs of the controllers do not change in the former one.
  
In order to limit this problem, we propose to regularize the schedules in such a way that the outputs of the controllers are stable (do not change too much between time steps). This can be achieved by extending the objective with a regularization term $S$ that sums the squares of changes in the charging speeds in consecutive time steps for each request 
$$S = \sum_{r \in \mathcal{R}} \frac{1}{L(r) - 1} \sum_{t, t-1 \in A(r)} (r_{t-1} - r_t)^2,$$
where $L(r)$ is the number of time steps the request $r$ is active and the rest of the variables is as above. The optimum solution for only this objective would be to use the same charging speed for the whole time the request is active. If we combine this with the original objective, we get a new optimization objective $$ \min \quad \mathrm{var_t}(B_t + \sum_{r \in \mathcal{R}\, s.t.\, t\in A(r)} r_t) + \lambda S, $$ where $S$ is the term above and $\lambda$ is the regularization parameter. We will call this type of regularization \emph{stabilization}.

\begin{figure}
    \centering
    \includegraphics[width=0.8\columnwidth]{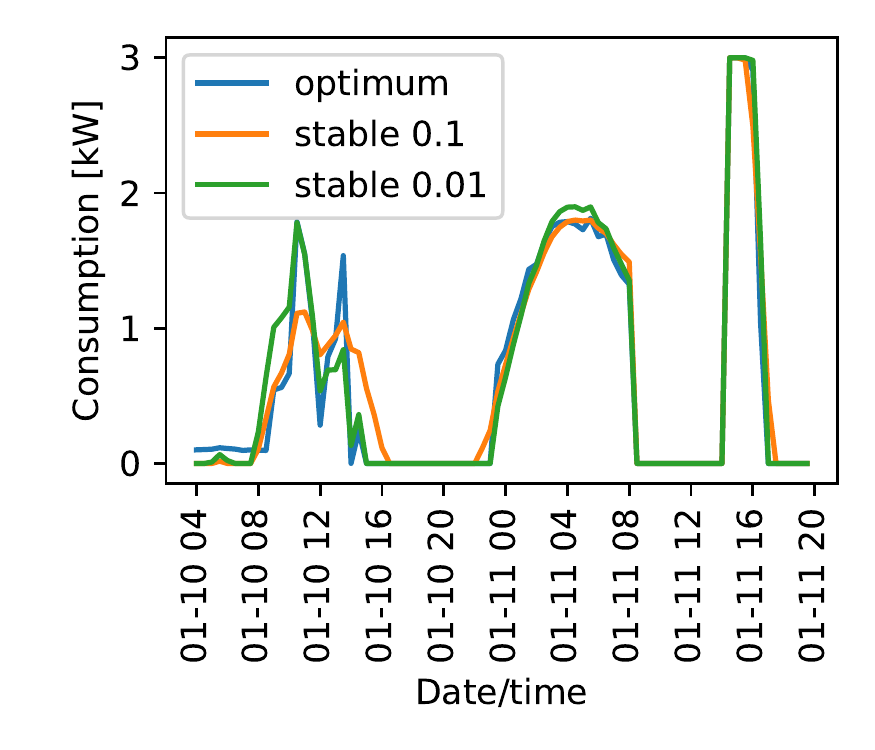}
    \caption{Comparison of the optimal charging schedule to those regularized for stable differences between consecutive time steps.}
    \label{fig:stabilization}
\end{figure}

The effect of the stabilization compared to the optimal schedule without the stabilization is demonstrated in Figure~\ref{fig:stabilization}. We can see that even for the parameter $\lambda = 0.01$ the solution is more stable, the change is even more pronounced for $\lambda = 0.1$, especially on the left side of the plot. We believe the more stable schedules should be easier to imitate for the controllers. Moreover, the stabilization, even with the small values of the parameter, should make the optimal solution less dependent on the specific solver used to find it.

\section{Experiments}
\label{sec:experiments}

In order to evaluate the performance of the trained controllers we ran a series of experiments. The experiments have three main goals:
\begin{enumerate}
    \item[G1] compare the imitation-based controllers to those created by direct optimization of the objective;
    \item[G2] evaluate, how close is the performance of the controllers to the one of the optimal schedules; and
    \item[G3] investigate the effect of the stabilization of the optimal schedules on the performance of the trained controllers.
\end{enumerate}

The data and source codes used in the experiments are available in a GitHub repository\footnote{https://github.com/martinpilat/evcharging/}.

\subsection{Data}

We need two types of data to perform the experiments -- information about the electricity consumption in a number of households and the charging requests that are based on the travel behavior of the users. In an ideal case, we would have both the consumption and the behavior from the same source, however, such a data set is not available at the moment. Therefore, we obtain each type of data from another source. In this paper, we use the electricity consumption data from UK Power Networks available as SmartMeter Energy Consumption Data in London Households in the London Datastore\footnote{https://data.london.gov.uk/dataset/smartmeter-energy-use-data-in-london-households}, and the user behavior from the National Household Travel Survey~\cite{nhts}. 

The electricity dataset contains electricity consumption measurements in 30 minute intervals from more than 5,500 houses in London. For the experiments we use data for the year 2013 and we selected 500 households from the dataset (the first 500 that do not contain many missing values -- the exact list of households is available in the source code repository). From these 500 households, we randomly selected 50 -- for these we generated the charging requests and simulated charging of an electric vehicle owned by them. We computed the average household electricity consumption of the rest 450 households in each time step and use it to simulate the consumption of households without EVs. Specifically, we multiply the average by 50 to simulate a baseline consumption of another 50 households without any electric vehicle. Obviously, this parameter can be set to other values in order to simulate different amounts of electric vehicles or different frequency of using the electric vehicles. Overall, we simulate a grid of 100 households of which 50 use an electric vehicle and a controller for its charging.

The data in the National Household Travel Survey contains the trips made by US households during a week. For each trip, information about time and distance is available together with the mean of transport used. From these, we filter the trips made by cars, and for each household we select the first trip of each day by car (starting at home) and the last trip of the day by car (going home). We assume the car is not at home (and therefore not available for charging) between these two times. We also compute the sum of the distances of all the car trips during the day (this is later used to compute the needed charge). Next, for every day and every household in the simulation we generate the charging requests. To this end, we first select for each day a random trip made in the same day of week and obtained as described above. The charging request then starts after the car comes home for the day and ends before the car leaves the next day. The amount of charge that needs to be charged is computed based on the distance travelled on the previous day. We compiled information on electricity consumption and battery capacities of seven different common types of electric vehicles. One of these is selected randomly and the needed charge is computed based on the distance and the consumption. In case the needed charge is larger than the battery capacity, we generate a random charge amount between 0 and the battery capacity (for such trips the battery must be charged somewhere during the day, therefore the charge needed during night is impossible to obtain). We also investigated the common charging speeds and in the end used a maximum charging speed of 3 kW for all the requests (there are faster chargers, however these are rarely used in households). Finally, we check if each of the charging requests can be theoretically fulfilled, i.e. if the battery can be fully charged using the maximum speed for the whole time. If it cannot, we remove the request and extend the previous one until the end of the removed one. Removing these requests is mostly technical, otherwise the quadratic problems would have no feasible solution. The requests also have no real effect on the comparison of the controllers, as all of them would use the maximum charging speed for the whole duration of the request and, additionally, only around 0.6 percent of the requests are removed this way. 

We consider the data created according to the steps above quite realistic -- we use real-world electricity consumption and travel habits. On the other hand, the data still have some limitations. First of all, the consumption and travels are from two geographically distant regions. Also the assumption that exactly half of the households use electric vehicles is not realistic, however, this number is merely a parameter in the experiments and further investigation for different values of this parameter is left as a future work.

Any training described below is always performed on the data from January 2013 and the results are presented on testing data from March 2013. In the computation of all the values, we always ignore the first 24 hours as these are used mainly to set the internal state of the controllers.

\subsection{Baseline Algorithms}

We use a number of baseline algorithms for comparison and in order to provide some reference for the values in the tables. These can be divided into 3 groups -- simple heuristic controllers, ideal schedules found by solving the quadratic problems, and controllers based on previous work~\cite{8489027}.

We show the results of three simple heuristic controllers. The MAX controller always uses the maximum possible charging speed until the whole request is fulfilled. Such a controller also simulates the situation where no controller is used and the car starts charging immediately when its connected to the charger. This typically causes large peaks in the afternoon when most people return home. Another simple controller is MIN. It uses the minimum required charging speed to still finish the charging in time -- this essentially means the cars are charged as late as possible and practically charges most of the cars at night. Finally, the last simple controller -- CONST -- uses the same (and thus lower) charging current for the whole duration of the charging request. 

The ideal controllers are based on solving the quadratic optimization problems defined in Section~\ref{sec:quad_opt}. The charging schedules based on all information without regularization (denoted by QP-opt) cannot be improved upon -- it is the global optimum of the problem. We also cannot expect to reach the same values with any of the controllers, as in the quadratic problem we assume full knowledge of future consumptions and future charging requests; such information is not available to the controllers. The regularized schedules (QP-st-0.1 and QP-st-0.01 for $\lambda$ set to 0.1 and 0.01 respectively) are presented to show the effect of regularization on the resulting schedules. We also show the schedule found, in case each household is considered separately (QP-h) to demonstrate that the controllers trained on all households at the same time can coordinate the charging better. 

As a more sophisticated baselines we use some of the models from our previous work~\cite{8489027}. As we use different dataset in this work compared to the previous one, we re-trained the models with the same parameters as in the original paper. Therefore we have controllers represented as neural networks with 5 units in the single hidden layer and one output neuron. We use ReLU as the activation in the hidden layer and logistic sigmoid as the activation in the output layer. The controllers are trained to minimize the standard deviation of the consumptions over time, either by the CMA evolution strategy~\cite{hansen2001completely} (population size of 16 individuals run for 250 generations) or by gradient descent using numerically computed gradients, i.e. each gradient computation means simulating the charging of EVs once for every parameter of the model (at most 250 iterations, step size is selected as the best one from 8 different values in each iteration). Both the CMA-ES and the gradient method generate the initial solutions as vectors from the normal distribution with standard deviation of 0.1. In the following we will refer to these controllers by names that consist of two parts -- the first part is the type of the controllers -- a single letter denoting if only the information from the household was used (H), or if also the information about the total consumption of the grid was provided (A). The other part of the name denotes the optimization algorithm used -- either the CMA-ES (denoted as CMA), or gradient descend (GD). In the previous paper, we did not use the information about the time when the request ends. The controllers that do not use this information (and therefore are exactly as those in~\cite{8489027}) are distinguished by appending NT (no time) to their name.

\subsection{Proposed Methods}

For the training of the models presented in this paper we first found the optimal solution to the respective quadratic problem for the months of January to March 2013 using the cvxpy package and the OSQP solver~\cite{osqp}. Then we generated the training set based on the January data as described in Section~\ref{sec:quad_opt}. For the training, the January data are further divided into training and validation sets in such a way that the first 865 data-points (roughly 85\%) for each household are used for training and the rest is used for validation. 

The controllers $\mathcal{C}(f, \theta)$ are represented as neural networks. We use two different architectures of the networks. Both have one hidden layer, the smaller one contains 5 neurons in the hidden layer, while the larger one contains 50 neurons. The output layer contains a single neuron that outputs a value between 0 and 1 that expresses the charging speed that should be used in the next time step. In all cases we use ReLU as the activation in the hidden layer and logistic sigmoid as the activation of the output layer. We also experimented with other activations, and higher numbers of neurons in the hidden layers, but we did not observe any significant differences in the MSE on the validation set. 

For the training, we implemented the models in tensorflow and we use the ADAM optimizer (batch size of 128 for at most 1,000 epochs) with early stopping (patience of 100 epochs) and we use the model from the epoch with the lowest mean square error on the validation set. After the training, the weights of the trained models are exported as a single long vector so that we can use the same implementation of controllers that is used by the EA or GD based baselines.

The various types of controllers are again distinguished by names combined from various parts. The first part is again a single letter that denotes the inputs that were used (H or A for household only and with grid consumption respectively). The next part denotes which schedule is imitated -- either the globally optimal one (OPT), or those with stabilization of the outputs (ST-0.1 and ST-0.01, where the number denotes the stabilization parameter $\lambda$). Finally, the last part denotes the number of neurons in the hidden layer (5 or 50). All these controllers also use the information about the time when the request ends.

\subsection{Results}

\begin{table*}
    \caption{The results of the baseline algorithms when evaluated on the data from March 2013. The top part of the table shows the results of the quadratic programming solvers using even the future information. The bottom part shows the results of the simple heuristic controllers. The objective is the standard deviation of the consumptions, the rest of the values are the consumptions in kW.}
    \label{tab:static}
    \centering
    \begin{tabular}{l|rrrrr}
        \hline
        {}         &  Objective &  Min load &   2.5\% &   97.5\% &  Max load \\
        \hline
        QP-opt     &       7.98 &     53.15 &   65.65 &    96.70 &    110.10 \\
        QP-st-0.01 &       7.99 &     52.58 &   65.03 &    96.71 &    110.10 \\
        QP-st-0.1  &       8.16 &     51.66 &   62.92 &    97.18 &    110.10 \\
        QP-h       &      14.81 &     44.08 &   52.13 &   107.26 &    117.25 \\
        \hline
        MAX        &      31.47 &     25.64 &   30.74 &   139.45 &    168.74 \\
        MIN        &      18.49 &     35.17 &   42.18 &   111.23 &    128.66 \\
        CONST      &      18.47 &     43.94 &   49.25 &   114.63 &    126.50 \\
        \hline
    \end{tabular}
\end{table*}

The results are summarized in Tables~\ref{tab:static} and \ref{tab:trained}. In both the tables, the "objective" column contains the standard deviation of the electricity consumption on the testing data (the month of March 2013 with the first 24 hours ignored to set the state of the controllers). The other columns than contain the statistics of the consumption -- minimum and maximum together with the 2.5-th percentile and 97.5-th percentile. We include the percentiles as the minima and maxima are often reached in only rare cases and the percentiles better represent the typical behavior.

In Table~\ref{tab:static}, we show the results of the deterministic controllers (MIN, MAX, CONST) and the results of the optimal charging schedules found by the quadratic optimization. In Table~\ref{tab:trained}, we show the results of the trainable controllers -- both for the baseline ones based on previous work~\cite{8489027} and the proposed ones. In these cases, the training can return a different controller each time due to random initialization or stochastic nature of the training, therefore we repeated all the experiments 10 times and report the average value and the standard deviation for the observed metrics.

The simple heuristic baselines do not perform very well. The MAX controller has the worst performance as it charges the cars as soon as they come home, which also corresponds to peaks in the baseline consumption. Interestingly, the MIN and CONST controllers perform almost the same as measured by the standard deviation of the overall consumption, but they differ greatly in the minimum load with the MIN controller dropping as low as 35 kW, while the CONST controller drops only to 44 kW. The minimum of the CONST controller is actually higher than the 2.5-percentile of the MIN controller.

\begin{table*}
    \centering
    \caption{The results of the trained controllers. The first part of the name denotes the inputs -- A for all and H for only the household ones. CMA and GD refer to the controllers trained by CMA-ES and gradient descend on the objective directly (as described in~\cite{8489027}). T denotes that the time of the end of request was used as a feature. OPT denotes controllers trained to imitate the optimal schedule, while ST denotes those trained to imitate the stabilized schedule with the number denoting the stabilization constant. The last number denotes the number of neurons in the hidden layer.}
    \label{tab:trained}
    \begin{tabular}{lrrrrr}
        \hline
        {} &     Objective &      Min load [kW] &        2.5 \% [kW] &       97.5 \% [kW] &       Max load [kW] \\
        \hline
        A-CMA-NT     &  $10.38\pm0.15$ &  $46.21\pm0.96$ &  $56.44\pm0.54$ &  $98.55\pm0.32$ &  $112.51\pm0.96$ \\
        A-CMA        &  $10.16\pm0.23$ &  $46.28\pm1.12$ &  $57.05\pm0.66$ &  $97.84\pm0.16$ &  $112.71\pm0.97$ \\
        A-GD-NT      &  $11.77\pm0.27$ &  $44.37\pm0.92$ &  $52.86\pm0.64$ &  $99.54\pm0.19$ &  $113.26\pm0.09$ \\
        A-GD       &  $11.41\pm0.41$ &  $44.61\pm1.18$ &  $53.47\pm1.01$ &  $98.66\pm0.37$ &  $113.19\pm0.07$ \\
        \hline
        A-OPT-5      &   $9.93\pm0.37$ &  $49.94\pm1.38$ &  $58.01\pm0.88$ &  $97.92\pm0.37$ &  $114.98\pm8.19$ \\
        A-OPT-50     &   $9.64\pm0.15$ &  $50.74\pm0.98$ &  $58.84\pm0.30$ &  $97.70\pm0.11$ &  $112.42\pm1.42$ \\
        A-ST-0.1-5   &  $10.31\pm0.40$ &  $48.34\pm1.51$ &  $56.60\pm1.03$ &  $98.13\pm0.27$ &  $112.12\pm1.06$ \\
        A-ST-0.1-50  &   $9.76\pm0.21$ &  $50.38\pm0.72$ &  $57.91\pm0.40$ &  $97.62\pm0.22$ &  $113.70\pm8.36$ \\
        A-ST-0.01-5  &  $10.07\pm0.21$ &  $49.22\pm1.11$ &  $57.46\pm0.78$ &  $98.16\pm0.23$ &  $113.03\pm0.47$ \\
        A-ST-0.01-50 &   $9.56\pm0.07$ &  $50.29\pm0.78$ &  $58.48\pm0.43$ &  $97.59\pm0.09$ &  $111.25\pm0.82$ \\
        \hline
        H-CMA-NT     &  $11.31\pm0.29$ &  $44.10\pm1.33$ &  $53.77\pm0.82$ &  $97.98\pm0.19$ &  $112.81\pm0.95$ \\
        H-CMA        &  $10.83\pm0.28$ &  $45.21\pm1.96$ &  $54.88\pm0.86$ &  $97.78\pm0.15$ &  $111.43\pm1.46$ \\
        H-GD-NT      &  $12.27\pm0.23$ &  $41.97\pm1.00$ &  $51.32\pm0.82$ &  $98.70\pm0.37$ &  $113.43\pm0.33$ \\
        H-GD         &  $12.16\pm0.14$ &  $41.75\pm0.81$ &  $51.38\pm0.75$ &  $98.60\pm0.09$ &  $113.28\pm0.05$ \\
        \hline
        H-OPT-5      &  $10.74\pm0.15$ &  $47.25\pm0.70$ &  $55.84\pm0.36$ &  $98.23\pm0.25$ &  $112.95\pm0.43$ \\
        H-OPT-50     &  $10.02\pm0.13$ &  $48.89\pm1.24$ &  $56.97\pm0.29$ &  $97.60\pm0.06$ &  $110.84\pm0.63$ \\
        H-ST-0.1-5   &  $10.97\pm0.29$ &  $45.58\pm0.79$ &  $54.94\pm0.54$ &  $98.76\pm0.44$ &  $112.65\pm0.68$ \\
        H-ST-0.1-50  &  $10.11\pm0.07$ &  $47.48\pm0.32$ &  $55.91\pm0.29$ &  $97.56\pm0.04$ &  $110.18\pm0.06$ \\
        H-ST-0.01-5  &  $10.88\pm0.22$ &  $46.61\pm0.86$ &  $55.28\pm0.66$ &  $98.49\pm0.25$ &  $112.71\pm0.41$ \\
        H-ST-0.01-50 &  $10.04\pm0.11$ &  $48.15\pm0.81$ &  $56.44\pm0.36$ &  $97.54\pm0.06$ &  $110.15\pm0.06$ \\
        \hline
        \end{tabular}
\end{table*}

The results of the trainable controllers are summarized in Table~\ref{tab:trained}. We can immediately see a few general, mostly expected, trends:
\begin{itemize}
    \item adding the information about the time when the request ends improves the performance of the controller in all cases,
    \item the controllers that use the information about the consumption of the whole grid generally outperform those that use only information about the household consumption,
    \item and controllers with larger network size are better than those with smaller networks.
\end{itemize}

We can also see that the controllers trained using the gradient descend on the objective are not better than those trained by the CMA-ES algorithm. This corresponds to the results from previous work~\cite{8489027}. 

\subsubsection*{EA/GD vs. imitation learning}

In most cases the controllers proposed here and trained by imitation learning are better than all the baselines. There are a few exceptions -- for example the A-ST-0.1-5 controller is slightly worse than the A-CMA controller, and the same holds for H-ST-0.1-5 and H-ST-0.01-5 compared to H-CMA. Generally, the controllers with the larger network are better than all the baselines. The best of these is the A-ST-0.01-50 that not only has the smallest values of the objective but also the smallest standard deviation in the performance between the different runs. The consumption of the whole grid when charging of EVs is controlled by this controller stays between 50.29 kW and 111.25 kW compared to 46.28 kW and 112.71 kW for the best baseline controller.

A similar observation can also be made for the controllers that use only the household information. The best of these -- H-OPT-50 keeps the consumption between 48.89 kW and 110.84 kW compared to the best baseline with 45.21 kW to 111.43 kW. Interestingly, the maximum consumption for the household-only controllers tends to be better than the maximum consumption for the controllers that have also the information for the whole grid. Specifically, the best controller in this regard is H-ST-0.01-50, and its maximum consumption of 110.15 kW comparable to the one of the optimal QP-opt schedule. On the other hand, even the worst of the trained controllers in this metric (A-OPT-5) is quite close to the optimal maximum with 114.98 kW. Interestingly, some of the controllers trained by imitation learning and using only the household data can beat even the baseline controllers using the data about the grid consumption (cf. H-OPT-50 and H-ST-0.01-50 vs. A-CMA).

Apart from the often superior performance, there is another advantage of training the controllers with imitation learning -- the whole training time, including solving the quadratic optimization problem, takes only minutes. The training with CMA takes around 1.5 hours and the gradient based training is even slower as it requires more simulations. This would make it possible to train the model multiple times using the same computational budget and selecting the best-performing one on a validation set. This would further increase the differences between the proposed method and the baselines.

\subsubsection*{Controllers vs. optimal schedules}

We already mentioned that regarding the maximum consumption, most of the controllers are quite close to the optimum. The same holds for the 97.5-th percentile. There are slightly larger differences in the minimum consumption, as for the optimal schedule it is 53.15 kW with the 2.5-th percentile being 65.65 kW. For the better baseline controllers, the minimum is around 46 kW (56 kW for the 2.5-th percentile) and for the ones proposed here, the minimum is around 50 kW with the 2.5-th percentile being around 58 kW. This shows that the proposed method performs better at times with lower consumption.

\subsubsection{The effect of regularization}

Comparing the results of the quadratic optimizers between themselves (cf. Table~\ref{tab:static}), we can see that the regularization of the outputs with the smaller parameter ($\lambda = 0.01$) does not have any significant impact on the performance. For the larger value of the parameter ($\lambda = 0.1$), the performance as measured by the objective is slightly worse, but there still do not seem to be any practically significant differences. In case we consider each household separately, the obtained schedule is much worse -- this makes sense, as the optimizer has no way to coordinate the charging. 

During the training of the controllers we (rather informally) observed the MSE of the networks on the validation set and, in the case with $\lambda = 0.1$, this metric was much smaller than in the case with no stabilization (roughly 0.002 vs 0.004). This indicates that the networks can indeed imitate the stabilized schedules better. On the other hand, when the controllers are evaluated in the simulation, their performance is not that different from the ones trained directly to imitate the optimal schedule. For the $\lambda = 0.1$ case, even though the predictions match the schedule better, the fact that the schedule itself is worse seems to have stronger effect and the overall result is a slightly decreased performance compared to the same model trained to imitate the optimal schedule. With the smaller stabilization parameter $\lambda = 0.01$ the performance is actually slightly better for the A-ST-0.01-50 controller than for the A-OPT-50 one with smaller differences between the runs. We therefore believe, there is further potential in these stabilization experiments and we will investigate them further in the future.

\begin{figure*}
    \centering
    \includegraphics[width=0.80\textwidth]{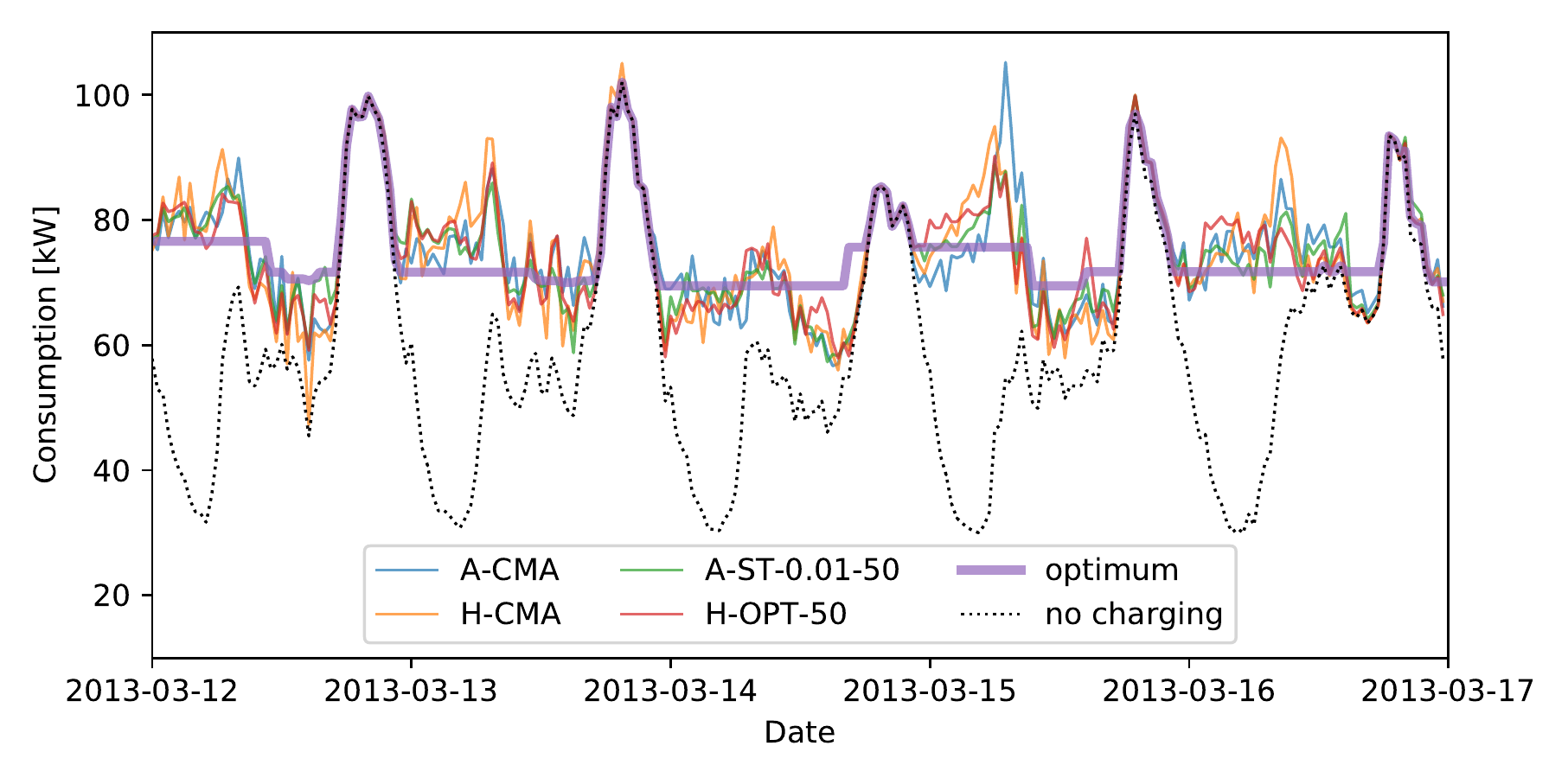}
    \caption{Comparison of the electricity load profiles while using different controllers for the charging.}
    \label{fig:loads}
\end{figure*}

In order to better visualize the differences in performance between the different controllers, we also include Figure~\ref{fig:loads}. It shows the optimal charging schedule found by the quadratic optimization together with the performance of the best controllers in each of the groups from Table~\ref{tab:trained}. We show the performance of the best controller found in the 10 runs. We can see that visually the differences are not very large and all the controllers are quite close to the optimum. However, both the EA-trained controllers (A-CMA, H-CMA) lead to higher peaks and deeper valleys than the best imitation-based controllers (A-ST-0.01-50 and H-OPT-50), which translates to the better performance of the latter ones.

\section{Conclusions and Future Work}

We have proposed a new method of training controllers for the coordination of electric vehicle charging. The method is based on imitating an optimal schedule found by solving a quadratic optimization problem based on a relaxed version of the original problem -- in the relaxation, we assume the future is known. The controllers can then be learned by supervised learning techniques instead of relying on reinforcement learning ones. This leads to faster training times and better performance.

One important feature of the controllers is that they respect the privacy of its users -- the charging requests are not shared with any third party, the controllers only require the information about the current consumption of the whole grid. We also investigated controllers based only on information available locally in the household.

From a practical point of view, the H-type controllers require no communication outside the household and can thus be deployed without any changes to the grid infrastructure. All the controllers presented here are also computationally efficient as they are only a small neural network (with at most 50 neurons in hidden layer) and can thus be easily run even on cheap embedded systems.

We believe the performance of the controllers can be further improved and the controllers themselves can be generalized. One of the limitations of the present work is that we assume (during training) a specific amount of households with electric vehicles (or more precisely with active charging requests). In the future, we would like to investigate how sensitive the controllers are to changes in this number and we would also like to extend the controllers so that they do not depend on this assumption.

Another possible future research direction is the regularization of the schedules found by the quadratic optimization. Here, we only tried stabilizing the changes in the charging speeds in consecutive time steps and we have shown that such stabilization can have advantages. We will investigate other possible types of regularization with the goal to make the training of the optimal schedules easier for the models, while preserving the quality of the schedules.

Finally, the transformation of the problem to supervised learning allows us to use almost any regression model instead of neural networks. Experiments with other types of models are also left as future work.

\section*{Acknowledgement}

Computational resources were supplied by the project "e-Infrastruktura CZ" (e-INFRA LM2018140) provided within the program Projects of Large Research, Development and Innovations Infrastructures.

\bibliographystyle{IEEEtran}
\bibliography{biblio.bib}

\end{document}